\def\year{2020}\relax
\documentclass[letterpaper]{article} 
\usepackage{aaai20}  
\usepackage{times}  
\usepackage{helvet} 
\usepackage{courier}  
\usepackage[hyphens]{url}  
\usepackage{graphicx} 
\usepackage{graphicx}  
\newcommand{\citet}[1]{\citeauthor{#1} \shortcite{#1}} \newcommand{\citep}{\cite} 
\usepackage{amsfonts}
\usepackage{amssymb}
\usepackage{bm}
\usepackage{bbold}
\usepackage{dsfont}
\usepackage{csquotes}
\usepackage{relsize}

\title{Induction and Reference of Entities in a Visual Story}
\author{ \Large \textbf{Ruo-Ping Dong}$^{\textbf{*}}$,  \textbf{Khyathi Raghavi Chandu}$^{\textbf{*}}$, Alan W Black\\ \Large \\  Language Technologies Institute\\ Carnegie Mellon University\\
\texttt{\{ruopingd, kchandu, awb\}@ cs.cmu.edu }
}
\begin{document}

\maketitle

\begin{abstract}
We are enveloped by stories of visual interpretations in our everyday lives. The way we narrate a story often comprises of two stages, which are, forming a central mind map of entities and then weaving a story around them. A contributing factor to coherence is not just basing the story on these entities but also, referring to them using appropriate terms to avoid repetition. In this paper, we address these two stages of introducing the right entities at seemingly reasonable junctures and also referring them coherently in the context of visual storytelling. The building blocks of the central mind map, also known as entity skeleton are entity chains including nominal and coreference expressions. This entity skeleton is also represented in different levels of abstractions to compose a generalized frame to weave the story. We build upon an encoder-decoder framework to penalize the model when the decoded story does not adhere to this entity skeleton. We establish a strong baseline for skeleton informed generation and then extend this to have the capability of \textit{multitasking} by predicting the skeleton in addition to generating the story. Finally, we build upon this model and propose a \textit{glocal hierarchical attention model} that attends to the skeleton both at the sentence (local) and the story (global) levels. We observe that our proposed models outperform the baseline in terms of automatic evaluation metric, METEOR. 
We perform various analysis targeted to evaluate the performance of our task of enforcing the entity skeleton such as the number and diversity of the entities generated. We also conduct human evaluation from which it is concluded that the visual stories generated by our model are preferred 82\% of the times. In addition, we show that our glocal hierarchical attention model improves coherence by introducing more pronouns as required by the presence of nouns. 

\end{abstract}

\section{Introduction}

\begin{quote}
\small
\textit{``You're never going to kill storytelling because it’s built in the human plan. We come with it.''   - Margaret Atwood}
\end{quote}

\begin{table*}[t!]
\scriptsize
\centering
\begin{tabular}{l | l l l | l l l }
\hline
 Sentences from SIS & Surface  & Nominalized & Abstract & Surface  & Nominalized & Abstract \\
 \hline
The cake was amazing for this event! & None & [0, 0] & None & event & [1, 0] & other \\
The bride and groom were so happy. & The bride and groom & [1, 0] & person & None & [0, 0] & None \\
They kissed with such passion and force. & They & [1, 1] & person & None & [0, 0] & None \\
When their son arrived, he was already sleeping. & their & [1, 1] & person & None & [0, 0] & None\\
After the event, I took pictures of the guests. & None & [0, 0] & None & event & [1, 0] & other \\
\hline    
\end{tabular}
\caption{\small Examples of three forms of Entity-Coreference Schema Representation}
\label{tab:coref}
\end{table*}

Storytelling in the age of artificial intelligence is not supposed to be a built-in capability of humans alone. With the advancements in interacting with virtual agents, we are moving towards sharing the ability of narrating creative and coherent stories with machines as well. The evolution of storytelling spans from primordial ways of cave paintings and scriptures to contemporary ways of books and movies. In addition, stories are ubiquitously pervasive all around us in digital media. This encompasses multiple modalities, such as visual, audio and textual narratives. In this work, we address narrating a story from visual input, also known as \textit{visual story telling} \cite{huang2016visual}. Generating textual stories from a sequence of images has gained traction very recently \cite{gonzalez2018contextualize,hsu2018using,kim2018glac,lukin2018pipeline,peng2018towards,chandustory}. Stories can be perceived as revolving around characters \cite{martin2018event}, events/actions \cite{rishes2013generating,mostafazadeh2016caters,peng2018towards},  or theme \cite{gervas2004story}. Emulating a naturally generated story requires equipping machines to learn where to introduce entities, and more importantly, how to refer to them henceforth. 

The main task addressed in this paper is to introduce entities similar to how humans do and more importantly, referring them appropriately in subsequent usage. We perform this in two phases: (1) Entity Skeleton Extraction, and (2) Skeleton Informed Generation. Here, a skeleton is defined as a simple template comprising of the entities and their referring expressions extracted using off-the-shelf NLP tools. These skeletons are extracted in three levels of abstraction, comprising of (1) surface form, the skeleton terms in the raw form, (2) nominalized form, that is the presence of entities in the noun or pronoun form, and (3) abstract form, that is using different notations for based on categories of words from language ontologies. This is delved in more detail in Section 4. We apply this for the task of visual storytelling which has both image captions and story sentences in a sequence. Leveraging the captions, the models also inherently learn the association of skeleton to the image captions thereby learning where to talk about which entities in a sequence of images.
Once this extraction is performed, we move on to the second phase of incorporating these coreference chains as skeletons to generate a story. 
The first approach is an incremental improvement over the baseline that performs multitasking with an auxiliary goal of predicting the entity skeletons to move the sequences generated from the primary task of story generation closer towards the extracted entities. The second approach is hierarchically attending to the entity skeletons at a local (corresponding to words within a sentence) and global (corresponding to the sentences making up the entire story) levels.

\section{Related Work}
 
\noindent \textbf{Visual Storytelling: }
\citet{huang2016visual}   proposed the first sequential vision-to-language dataset, comprising of sequences of story-like images with corresponding textual descriptions in isolation and stories in sequences. \citet{kim2018glac} proposed a seq2seq framework and \citet{smilevski2018stories} proposed late fusion techniques to address this task. 
We derive motivation from these techniques to introduce entities and references as skeletons. \citet{park2015expressing,liu2017let} explored the task of generating a sequence of sentences for an image stream. 
\citet{agrawal2016sort} introduced the task of sorting a temporally jumbled set of image-caption pairs from a story such that the output sequence forms a coherent story. \citet{liu2017let} proposed a joint embedding of photos with corresponding contextual sentences that leverages the semantic coherence in a photo sequence with a bidirectional attention-based recurrent model to generate stories from images. 

\noindent \textbf{Schema based generation: } 
\citet{grosz1995centering} was one of the initial works delving into how entities and their referring expressions are used in a discourse context. Several research efforts for narrative generation tasks have spawned from introducing a schema or a skeleton. \citet{martin2018event,clark2018neural} explored the usage of event representations and predicting successive event forms to generate the entire story. \citet{fan2018hierarchical} proposed hierarchical frameworks for story generation conditioned on a premise or a topic. This work was also extended by decomposing different parts of the model by generating a surface realization form of the predicate-argument structure by abstracting over entities and actions \cite{fan2019strategies}. 
\citet{xu2018skeleton} used reinforcement learning to first generate skeleton (the most critical phrases) and then expand the skeleton to a complete sentence. \citet{yao2018plan} proposed a hierarchical generation framework in which given a topic, the model first plans a storyline, and then generates a story based on the storyline. Recently, \citet{zhai2019hybrid} proposed a model to generate globally coherent stories from a fairly small corpus by using a symbolic text planning module to produce text plans, and then generating fluent text conditioned on the text plan by a neural surface realization module. \citet{ammanabrolu2019guided} showed that event-based generation often generated grammatically correct but semantically unrelated sentences and present ensemble methods for event based plot generation as a solution.

Our work falls along the lines of generating a story from visual input based on schema. However, to the best of our knowledge, this work is the first to combine the spaces of both generating a visual story from a skeleton schema of entities and how they are referred henceforth.


\section{Data Description}

A dataset that has recently gained traction in the domain of visual story telling is proposed by \citet{huang2016visual}. This problem of grounded sequential generation is introduced as a shared task\footnote{\small{\url{visionandlanguage.net/workshop2018/index.html\#challenge}}}.
Formally, the dataset comprises of visual stories $\bm{S} = \{\bm{S_1}, \ldots, \bm{S_n} \}$. Each story in the dataset consists of a sequence of five story-like images, along with descriptions-in-isolation (DII) and stories-in-sequences (SIS). The descriptions in isolation are isomorphous to image captions. Each story can be formally represented as $\bm{S_i} = \{(\bm{I}_i^{(1)}, \bm{x}_i^{(1)}, \bm{y}_i^{(1)}), \ldots, (\bm{I}_i^{(5)}, \bm{x}_i^{(5)}, \bm{y}_i^{(5)})\}$, where $\bm{I}_i^{(j)}$, $\bm{x}_i^{(j)}$ and $\bm{y}_i^{(j)}$ are each image, single sentence in DII and single sentence in SIS respectively, and \textit{i} refers to the \textit{ith} example story. SIS and DII are supposed to be associated with each image. However there are about 25\% of the images for which DII are absent in the dataset. The corresponding statistics of the dataset are presented in Table \ref{tab:data}.

\begin{table}[h]
\small
\centering
\begin{tabular}{l l l l }
\hline
 & Train & Val & Test \\
 \hline
\# Stories & 40,155 & 4,990 & 5,055 \\
\# Images & 200,775 & 24,950 & 25,275 \\
\# with no DII & 40,876 & 4,973 & 5,195 \\
\hline    
\end{tabular}
\caption{\small Details of the Dataset}
\label{tab:data}
\end{table}

In our modeling approaches as described in the next section, we also need the descriptions in isolation.
Hence for the images for which the DII are absent, we use a pre-trained image captioning model to make the dataset complete for our use case.


\begin{figure*}[t!]
\centering
\includegraphics[width=0.49\linewidth]{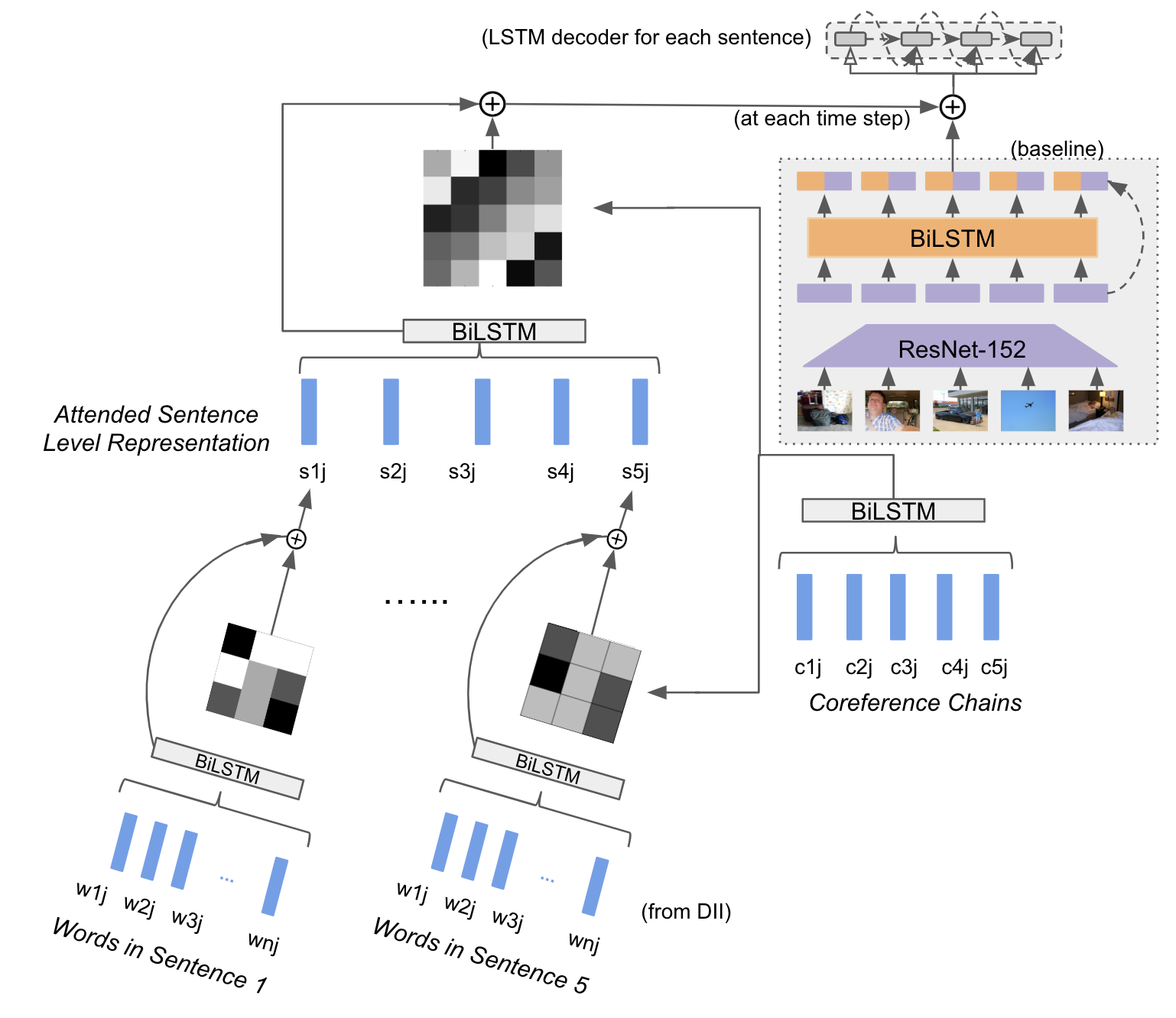}
\caption{ {\small Architecture of Glocal Hierarchical Attention on Entity skeleton coreference chains to perform Visual Storytelling} }
\label{fig:coref_arch}
\end{figure*}

\section{Model Description}
\label{sec:models}

Our approach of using entity skeletons to generate a coherent visual story is divided into two phases: (1) Entity Skeleton Extraction, and (2) Skeleton Informed Generation. In this section, we first describe 3 kinds of schema extraction for coreference chains and then proceed towards describing two baselines and two proposed story generation models. We will be releasing the entire streamlined codebase.


\subsection{Entity Skeleton Extraction}

The task is to introduce the characters in right times and refer to them appropriately henceforth. This means that we not only target the head mention of an entity but also cater to the corresponding appropriate coreference expressions. We define the skeleton as a linear chain of entities and their corresponding referring expressions. There could be multiple coreference chains in a long narrative. We associate a story with the entity skeleton that has maximum representation in the five sentences. This means that the elements of the skeleton need to be present in the majority of the sentences, thus making it the central theme for basing the story on. For simplicity purposes, in case of a tie with respect to the above criterion of the number of mentions, we select the skeleton to be the most frequently occurring coreference chain from among all the chains that are present in it. In our future work, we plan to extend this capability to cater to multiple skeletons simultaneously. We use off-the-shelf tools to represent these skeletons in three different ways. 

For each of the following skeleton representations, we first extract the coreference chains from the textual stories that are made up of SIS in the training data. This is done by using version 3.7.0 of Stanford CoreNLP toolkit \cite{manning2014stanford}. These three ways of representing skeletons are described in detail next.

\subsubsection{1. Surface form Coreference Chains: }
The resulting coreference chains now comprise of surface word forms of entities and their corresponding reference expressions. In specific, the skeleton for each story is represented as $\{\bm{c}_1, \ldots, \bm{c}_5\}$, where $\bm{c}_j$ is the coreference word in \textit{jth} sentence. An example of this can be seen in Table \ref{tab:coref}. From the story sentences on the left, there are two entity chains that are extracted corresponding to \textit{`the bride and the groom'} and \textit{`event'}. The skeleton word is \textit{None} when there is no word corresponding to that coreference chain in that sentence. The surface form entity skeletons for each sentence are shown in the following two columns. Note that there could be multiple such chains extracted for each story owing to the number of different entities that are present in the story. Our goal is to pivot the story on a central mind map, so we select the chain that has minimum number of \textit{Nones} in the five sentences. Hence in this example, we go ahead with the first skeleton with \textit{`the bride and groom'} to weave the story since the skeleton with \textit{`event'} has higher number of \textit{Nones}.

\subsubsection{2. Nominalized Coreference Chains: } The surface form entity skeletons extracted as described before do not comprise the information of whether it is the head mention of the entity or whether it is referred later. In crude terms, it does not cater to abstracting the properties of the skeleton words from the surface form word itself. The remaining two forms of skeleton representations address this issue of abstracting the lexicon from the properties of the word. In order to encode this information explicitly, we disintegrate the bits that correspond to the properties of presence and absence of the entity words and whether the word is present in the noun or the pronoun form. The skeleton for each story is represented as  $\{ [h, p]_{1}, \ldots, [h, p]_5\}$. Here, \textit{h} $\in$ \{0,1\}, is a binary variable indicating if there is a coreference mention, i.e 1 if there is a mention in the skeleton chain and 0 if it is None. Similarly, \textit{p} $\in$ \{0,1\} is a binary variable indicating that the word is head mention i.e, the word is in the noun form if it is 0 and pronoun form if it is 1. For instance, in Table \ref{tab:coref}, in sentence 2, the skeleton is represented as [1,0] which means that this sentence has a mention of the skeleton under consideration and it is in the noun form. Note that we do not use the surface representation of the word itself while we represent the skeleton in this format.

\subsubsection{3. Abstract Coreference Chains: } As observed from Table \ref{tab:coref}, the skeleton chains belong to different categories of entities. Instead of disintegrating the properties into noun and pronoun, another form is to represent them into the abstract categories that they belong to. These categories can be \textit{person, object}, \textit{location} etc., This differentiates the order of introduction and references of objects or people in the timeline among the five sentences. We use Wordnet \cite{miller1995wordnet} to derive these properties. As depicted in Table \ref{tab:coref}, the entity skeleton corresponding to a coreference chain can be represented with a sequence of \textit{`person', `other'} and \textit{`None'}.

\subsection{Schema Informed Generation}

In this section, we describe the baseline model used to generate textual stories from visual input. 
In order to establish a fair comparison, we alter this baseline slightly to establish a second baseline that accesses the skeleton information. We then move onto discussing two models that incorporate the entity skeletons that are described in the previous section.

\begin{table*}[t!]
\small
\centering
\begin{tabular}{l l l l l}
\hline
\textbf{Models} &\textbf{Entity Skeleton Form} & \textbf{METEOR} & \textbf{Distance} & \textbf{Avg \# distinct entities} \\
\hline
Baseline & None & 27.93 & 1.02 & 0.4971 \\
Baseline with Entity Skeletons & Surface & 27.66 & 1.02 & 0.5014 \\
MTG ($\alpha$(0.5)) & Surface & 27.44 & 1.02 & 0.9554\\
MTG ($\alpha$(0.4)) & Surface & 27.59 & 1.02 & 1.1013 \\
MTG ($\alpha$(0.2)) & Surface & 27.54 & 1.01 & 0.9989 \\
MTG ($\alpha$(0.5)) & Nominalization & \textbf{30.52} & 1.12 & 0.5545 \\
MTG ($\alpha$(0.5)) & Abstract & 27.67 & 1.01 & 0.5115 \\
Glocal Attention & Surface & \textbf{28.93} & 1.01 & \textbf{0.8963} \\
\hline    
\end{tabular}
\caption{Automatic Evaluation of Story Generation Models}
\label{tab:alpha}
\end{table*}

\subsubsection{1. Baseline Model: }

\noindent Our baseline model has an encoder-decoder framework that is based on the best performing model in the Visual Story Telling challenge in 2018 \cite{kim2018glac} that attained better scores on human evaluation metrics. The model essentially translates a sequence of images to a story. All of the images are first resized to 224 X 224 and image features are extracted from the penultimate layer of ResNet-152 \cite{he2016deep}. These image features act as local features for decoding the sentence corresponding to that image. This sequence of image features are passed through two layers of Bi-LSTMs in order to obtain the overall context of the story. This contributes to \textit{global} theme of the story. The \textit{local} context for each sentence in the story is incorporated with a skip connection of the local features for that particular image. The global and local features are concatenated and passed to each time step in the LSTM decoder to generate the story word by word. 

For simplicity in formal representation, we use the following notations. Subscript \textit{t} and superscript $\tau$ indicates the \textit{$t^{th}$} step or sentence in a story and \textit{$\tau^{th}$} word within the sentence respectively. 
\textit{$I_{t}$}, \textit{$x_{t}$}, \textit{$y_{t}$}, represent image, DII, SIS for a particular time step. \textit{$k_{t}$} is the  skeleton coreference element for that particular sentence. Here \textit{k} can take any of the three forms of coreference chains discussed previously, which is word itself (surface form) or a pair of binary digits (nominalization) or noun properties (abstract). Note that \textit{k} is not used in this baseline model. 

The encoder part of the model is represented as the following which comprises of two steps of deriving the local context features ${l}^t$ and the hidden state of the \textit{$t^{th}$} timestep of the BiLSTM that gives the global context. 

$\boldsymbol{l}_t =\textit{ResNet} (\boldsymbol{I}_t)$

$ \boldsymbol{g}_t = \textit{Bi-LSTM} ( [{l}_1, {l}_2 ... {l}_5 ]_t  )$

The latent representation obtained from this encoder is the \textit{glocal} representation $[\boldsymbol{l}_t, \boldsymbol{g}_t ]$, where [..] represents augmentation of the features. This \textit{glocal} vector is used to decode the sentence word by word. The generated words in a sentence from the decoder \boldsymbol{$\hat{w}_{t}$} is obtained from each of the words \boldsymbol{$\hat{w}^{\tau}$} that are the outputs that are also conditioned on the generated words so far \boldsymbol{$\hat{w}_t^{<{\tau}}$} with $\tau^{th}$ word in the sentence being generated at the current step.

\begin{equation}
    \boldsymbol{\hat{w}}_{t} \sim \prod_\tau Pr(\boldsymbol{\hat{w}}_t^{\tau} | \boldsymbol{\hat{w}}_t^{<{\tau}}, \boldsymbol{l}_t, \boldsymbol{g}_t)
\end{equation}

The baseline model is depicted in the right portion of the Figure \ref{fig:coref_arch}.

\paragraph{2. Skeleton Informed Baseline Model: } 

We need to make a note here that though the above baseline is the best performing model in the task, it does not take into account for the explicit mentions of the entities as a skeleton to weave the story on. Similarly, it does not make use of the DII for the images. We explore on how to make better use of these DII to extract the entity skeletons. Hence to establish a fair comparison with our proposed approaches we condition the decoder on not only the glocal features and the words generated so far, but also the surface form of the words.

\begin{equation}
    \boldsymbol{\hat{w}}_{t} \sim \prod_\tau Pr(\boldsymbol{\hat{w}}_t^{\tau} | \boldsymbol{\hat{w}}_t^{<{\tau}}, \boldsymbol{l}_t, \boldsymbol{g}_t, \boldsymbol{k}_t)
\end{equation}

In specific the features that are given to the decoder now have [$\boldsymbol{l}_t, \boldsymbol{g}_t, \boldsymbol{k}_t$]. The skeleton information is provided to every time step in the decoder.

\paragraph{3. Multitask Story Generation Model (MTG): } 

Incorporating the entity skeleton information directly in the decoder might affect the language model of the decoder. Hence we take an alternate approach that incrementally improves upon the first baseline model to enable it to perform two tasks. Instead of augmenting the model with skeleton information, we enable the model to predict the skeleton and penalize it accordingly. The main task here is the generation of the story itself and the auxiliary task is the prediction of the entity skeleton word per time step. Each of these tasks are optimized using cross entropy loss. The loss for generation of the story is $\textbf{L}_{1}$ and the loss to predict the skeleton of the model is $\textbf{L}_{2}$. However, we do not want to penalize the model equally for both the participating tasks and weigh them by a factor $\alpha$ as much as to affect the language model of the decoder. We experimented with different weighting factors for $\alpha$ which are presented in Table \ref{tab:alpha}.

\begin{equation}
\setlength\abovedisplayskip{-1pt}
\textstyle
\nonumber
\mathlarger { \sum\limits_{I_{t}, x_{t}, y_{t} \in \mathbb{S}} } \alpha \textbf{L}_{1}(I_{t}, y_{t}) + (1-\alpha) \textbf{L}_{2}(I_{t}, y_{t}, k_{t}) 
\label{eq:loss2}
\end{equation}

Note that we do not use \textit{k} as a part of the encoder even in this model but only use them to penalize the model when the decoded sentence does not contain skeleton similar to \textit{k}.

\paragraph{4. Glocal Hierarchical Attention: } 
Enabling the model to predict the entity skeleton equips it to model the sentences around the entities, thereby weaving the stories around the skeleton. However, this multitasking model does not explicitly capture the relationship or focus on the words within a sentence or across the five sentences with respect to the skeleton in consideration. Hence, we went one step further to identify the correlation between the coreference skeleton with different levels including within a sentence (i.e, at word level) and across sentences (i.e, at sentence level). We use attention mechanism to represent these correlations.

We propose two stages of attention to capture this information:
\begin{enumerate}
    \item \textit{Local Attention: }attending to the words in captions (\textit{$w_t^{\tau}$} from \textit{$x_t$}) with respect to the entity skeletons \textit{$k_{t}$}.
    \item \textit{Global Attention: }attending to the sentences in the story derived from the local attention for each sentence.
\end{enumerate}

Figure \ref{fig:coref_arch} depicts the entire glocal hierarchical attention model with the encoder decoder framework on the right and the two stages of attention on the left. The attention is performed on textual modality corresponding to DII ($x_t$) and hence can be perceived as translating DII to SIS. As observed in Table \ref{tab:data}, DIIs are absent for about 25\% of the data. We use an image captioning model pretrained on ImageNet data \cite{ILSVRC15}. These image captions are substituted in the place of missing DII.

\noindent \textbf{Local Attention: } The first level of attention i.e, the \textit{local attention} measures the correlation between words in each sentence to the coreference skeleton words. There are 5 sentences in each story corresponding to five images. Since we use the skeleton words as they appear to attend to the words in DII, we use the surface form notation in this model. As we have seen, the surface form skeleton is represented as $\textit{C} = \{c_1, c_2.., c_5 \}$. The vocabulary of these surface form skeleton words is limited to 50 words in the implementation. The surface skeleton form \textit{C} is passed through a Bi-LSTM resulting in hidden state $H_{k}$ which is of 1024 dimensions. This hidden state is used to perform attention on the input words of DII for each image. Note here that the skeleton words for coreference chains are extracted from SIS (i.e, from $\{y_1, y_2.., y_5 \}$), from which the hidden state is extracted, which is used to perform attention on the individual captions (DII i.e, $\{x_1, x_2.., x_5 \}$ ).

The skeleton remains the same for all the sentences. The skeleton form is passed through a Bi-LSTM resulting in  $H_{k} \in \mathbb{R}^{k  \times 2h}$, where hidden dimension of the Bi-LSTM is \textit{h}. Each $x$ in the story (with \textit{n} words in a batch) is passed through a Bi-LSTM with a hidden dimension of \textit{h}, resulting in $H_{w} \in \mathbb{R}^{5 \times n \times 2h}$. This then undergoes a non-linear transformation. 

Attention map for the word level is obtained by performing a batch matrix multiplication (represented by $\otimes$) between the hidden states of the words in a sentence and the hidden states of the entity skeleton. In order to scale the numbers in probability terms, we apply a softmax across the words of the sentence. Essentially, this indicates the contribution of each word in the sentence towards the entity skeleton that is present as a query in attention. This is the \textit{local attention} $A_{w} \in \mathbb{R}^{5 \times n \times k}$ pertaining to a sentence in the story. Mathematically, equation \ref{eq:attnword} depicts the calculation of \textit{local attention}.

\begin{equation}
\setlength\abovedisplayskip{-1pt}
\textstyle
\nonumber
A_{w} = softmax( H_{w} \otimes H_{k} )
\label{eq:attnword}
\end{equation}

\begin{figure*}[!tbp]
  \centering
  \begin{minipage}[b]{0.75\textwidth}
    \includegraphics[width=0.99\textwidth]{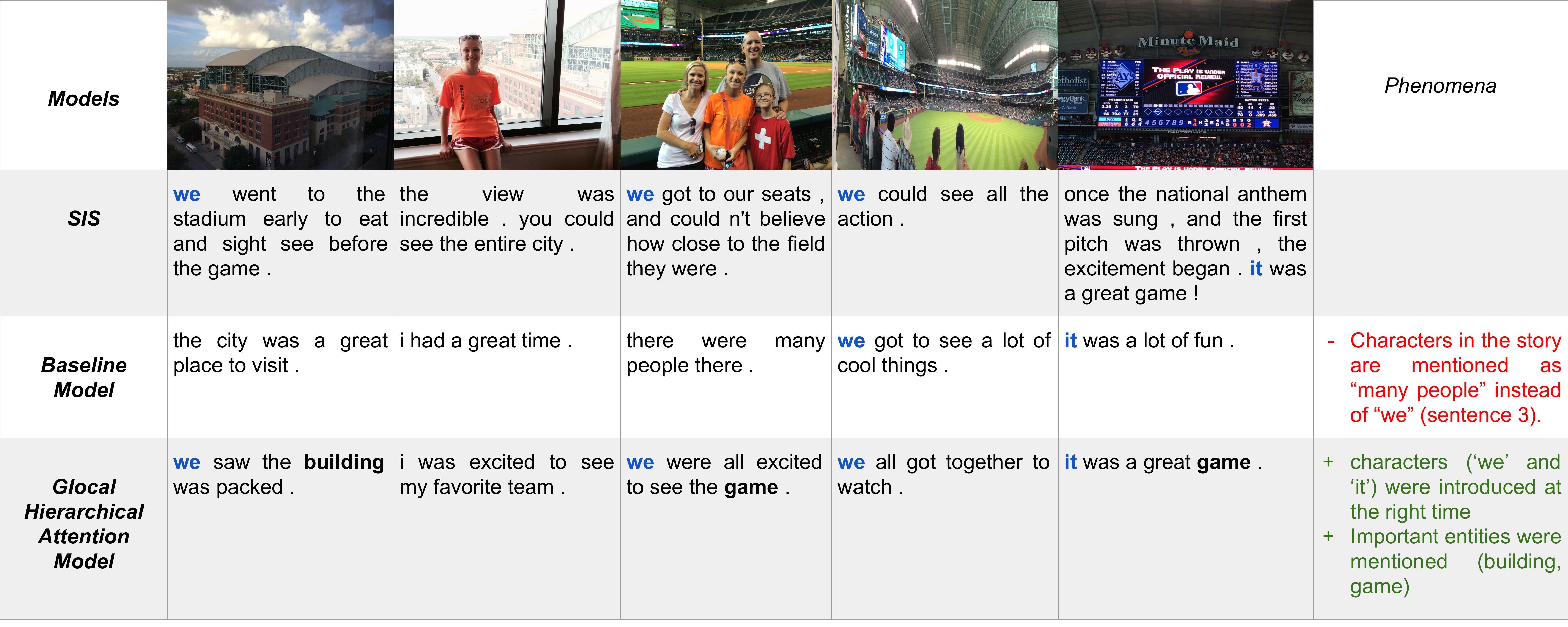}
    \caption{{\small Qualitative Analysis: Example of generated stories}}
    \label{fig:coreference_analysis}
  \end{minipage}
\end{figure*}

\begin{figure}[!tbp]
\centering
 \begin{minipage}[b]{0.29\textwidth}
    \includegraphics[width=0.99\textwidth]{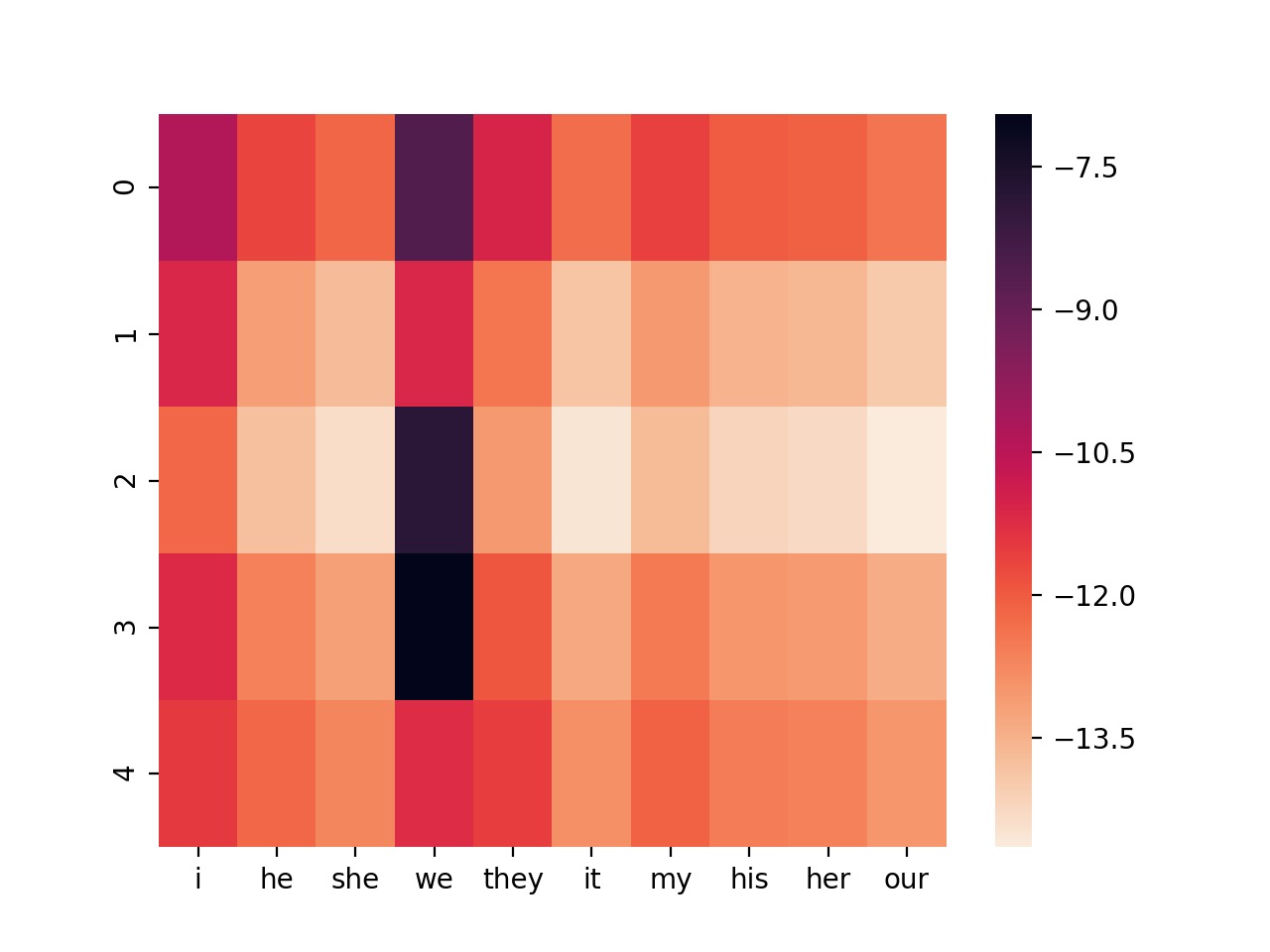}
    \caption{ {\small Visualization of the Glocally Attended representation of the skeleton for story in Figure 2}}
    \label{fig:attention_visualization}
  \end{minipage}
  \end{figure}

\noindent \textbf{Glocal Attention:} We then perform \textit{global attention}, which is at the entire story level. In other words, this attention evaluates the contribution of different sentences in the story towards responding to the entity skeleton that is extracted. Instead of considering the sentence representation as the output of passing the words as is through a Bi-LSTM, we leverage the already attended local attention (which is at sentence level) to perform the global attention. Hence it is the combination of global and local attention, and thereby we perform \textit{glocal hierarchical attention}.

For this, the locally attended representation of each sentence is then augmented with the output of the Bi-LSTM that takes in DII. The attended representation for each of the \textit{k} words are concatenated and projected through a linear layer into 256 dimensions ($P_{w}$). This goes in as sentence representation for each of the $s_{ij}$ (where \textit{i} is the index of the sentence in the story and \textit{j} corresponds to the story example) as shown in Figure \ref{fig:coref_arch}.
The word representations at each time step are obtained by augmenting the corresponding vectors from $H_{w}$ and $P_{w}$. These form our new sentence embeddings. These sentence embeddings are again passed through a Bi-LSTM to get a sentence level representation. This process is done for each sentence in the story (which are the replications as shown in the left portion of Figure \ref{fig:coref_arch}). This results in a latent representation of the story $H_{s} \in \mathbb{R}^{5 \times 2h}$. Along the same lines of local attention, we now compute story level hierarchical global attention to result in $A_{s} \in \mathbb{R}^{5 \times k}$. This is shown in Equation \ref{eq:attnstory} where $[,]$ indicates augmentation of corresponding vectors.

\begin{equation}
\setlength\abovedisplayskip{-1pt}
\textstyle
\nonumber
A_{s} = softmax( [H_{w}, P_{w} ] \otimes H_{k} )
\label{eq:attnstory}
\end{equation}

The attended vectors from $A_{w}$ and $A_{s}$ of size \textit{nk} and \textit{k} respectively are concatenated in each sentence step in the decoder from the baseline model. This is shown in the top right corner of Figure \ref{fig:coref_arch} (although the Figure depicts concatenation for single time step).

\noindent \textbf{Hyperparameter setup: } Learning rate of 0.001 is used with a batch size of 64. The word embedding dimension is 256 and the image features contributing towards the local representation is 1024. The hidden size of the Bi-LSTM is 1024 which is the dimension of the global vectors. The attention map features are of dimension size of 256.  We selected the 50 most frequently occurring coreference words to set the vocabulary of \textit{k} for these experiments i.e, $size(k)=50$.


\section{Experiments and Results}

This section presents the quantitative and the qualitative results for the four models discussed in the previous section. We first discuss the automatic evaluation and some qualitative analysis of our models.

We perform automatic evaluation with METEOR score for generation. The results are shown in Table \ref{tab:alpha}.
However, our main target is to verify whether the story adheres to the entity skeleton form that is provided. Hence we attempt to perform a different scoring mechanism.
We extract entity skeletons from the generated stories in the same procedure as performed on the training stories. With respect to the ground truth stories, a binary vector of length 5 is constructed based on whether the entity skeleton word is present or not in that sentence. Euclidean distance between these binary vectors skeletons of the original and the generated stories is used to validate this aspect of generation. Table \ref{tab:alpha} presents the results of our models. As we can see, the Euclidean Distance is not very different in each of the cases. However, we observe that the multitasking approach (MTG) is performing better with nominalization form of entity skeletons as compared to the baselines and other forms of entity skeleton representations as well. The \textit{glocal} model described performs attention on the surface words only and hence the experiment includes only this configuration. We observe that glocal attention model outperforms the baseline model. However, there is a scope for improvement when the attention mechanism is performed on nominalized skeleton representation, which we leave for the future work.

\begin{figure}[t!]
\centering
\includegraphics[width=0.9\linewidth]{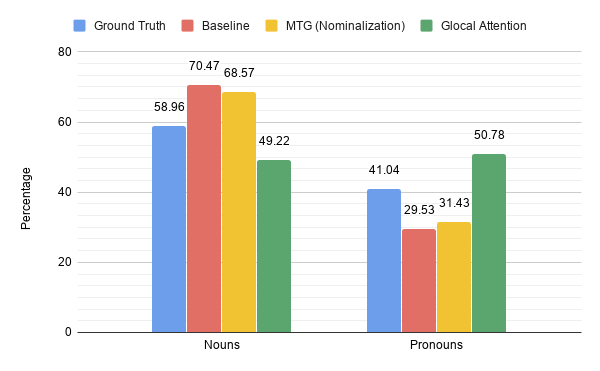}
\caption{ {\small Percentage of Entities in the form of Nouns and Pronouns in the generated stories } }
\label{fig:np_analysis}
\end{figure}

These automatic metrics do not sufficiently capture the number or the diversity of the entities that are introduced in the generated stories. For analyzing the number of entities, we calculated the percentages of the nouns and pronouns in the ground truth and the generated stories for the test data. Figure \ref{fig:np_analysis} presents these percentages for the ground truth stories, generated stories from baseline, MTG with nominalized skeleton representation and the Glocal attention model. As we can see on the nouns section, the baseline model seemed to have over-generated nouns in comparison to both of our proposed models. While our MTG model also has over-generated the nouns, our glocal attention model has generated fewer nouns compared to the ground truth. However, this is still the closest to the number of nouns in the ground truth stories. Generating high number of nouns does not ensure coherence as much as generating appropriate number of relevant pronouns. This is observed in the second section in the graph. While the MTG model generated higher number of pronouns in comparison to the baseline, the glocal attention model seemed to have generated even higher percentage of pronouns. Despite this over-generation, glocal attention model is the closest to the number of pronouns in the ground truth stories. An interesting observation is that the MTG and glocal attention models seem to have opposite trends in the generation of nouns and pronouns. We plan on investigating this further in our future work. 
Coming to the diversity of the entities generated by the stories, we calculate the average number of distinct entities present per story for each of the models. These numbers are shown in the last column of Table \ref{tab:alpha}. This number for the ground truth test stories is \textit{0.7944}. As we can see, the average number of distinct entities is comparatively high for the MTG model. However this number is closer to that of the ground truth for the glocal attention model assuring that there is sufficient diversity in the entity chains that are generated by this model.
We would like to make a note here that though MTG model with nominalized representation is performing better in terms of METEOR score, our analysis shows promisingly better performance of the Glocal attention model with respect to both the number and diversity of the entities generated.

\noindent \textbf{Qualitative Analysis:} Figure \ref{fig:coreference_analysis} presents an image sequence for a story along with the corresponding ground truth (SIS) and the generated stories. The positive and the negative phenomena observed are presented in the last column. The Glocal Hierarchical Attention Model is able to capture the skeleton words right in comparison to the baseline model. For instance, the words \textit{`we'} and \textit{`it'} are generated in sentences 1, 3, 4 and 5 with the glocal attention model whereas these entity skeleton words are generated only in sentences 4 and 5 in the baseline model. Since there are multiple occurrences of entities that are connected, the story might present stronger coherence in the case of glocal attention model. In addition, the entity skeleton could be boosting the model to also generate other relevant words based on images such as \textit{`building'} and \textit{`game'}. The visualization of the corresponding hierarchical glocal attention map that is fed into the decoder is presented in Figure \ref{fig:attention_visualization}. Darker color indicates higher attention on those words. The rows in the visualization depict the sentence indices and the columns indicate a few of the frequently occurring entity skeleton chains. The scores are not normalized as probability distributions since the figure does not present all 50 of the entity skeletons (instead of only the top 10 frequently occurring ones). As we can see, there is higher weight in the grids pertaining to \textit{`we'} for the first, third and fourth sentences.

\noindent \textbf{Human Evaluation:} We conduct human evaluation in the form of preference testing. 20 stories were randomly sampled and we asked five subjects the following preference questions \textit{`preference of the story narrative from the images'}. Our \textit{glocal hierarchical attention model} is preferred 82\% of the times compared to the baseline model and 64\% of the times in comparison to the MTG model with nominalized representation. We also asked them a follow up question of what their opinion is on the usage of pronouns since that is the task we were focusing on. From the answers, we conclude that our hypothesis of the usage of pronouns instead of third party nouns narrates a more involved story. Therefore, this provides an opportunity margin for improving story generation.


\section{Conclusion and Future Work}

Automatic storytelling has been a dream since the emergence of AI, with one of the main hurdles being naturalness. Naturalness to a story comes as a package of not only  introducing entities, but also referring to them appropriately as humans do. Our work is inspired from the intuition that humans form a central mindmap of a story before narrating it. In this work, this mindmap is associated with the entities (such as people, location etc.,) involved in the story.
We present our work on introducing entity and reference skeletons in the generation of a grounded story from visual input. 
In the first phase, we represent the entity skeletons in three forms: surface, nominalized and abstract. These forms of representations correspond to different properties of the skeleton words like whether they are nouns or pronouns or the category of the nouns based on an ontology. In the second phase, we present a story generation model that takes in the entity skeletons and the images. A strong baseline model is selected that has high performance with respect to human evaluation scores for the task of plain visual storytelling. We extended the baseline to setup another baseline that is informed of the entity skeletons to perform a fair comparison. We then proposed two models: (1) multitasking with the prediction of the skeleton, and (2) glocal hierarchical attention model that attends to the skeleton words at the word level and the sentence level hierarchically. We observe that our \textit{MTG} and \textit{glocal hierarchical attention} models are able to adhere to the skeleton thereby producing schema based stories. Our MTG model is performing better in terms of automatic metrics like METEOR by around 3 units. However, analysis on the percentage of generation of the noun and pronoun forms of entities reveals that the glocal hierarchical attention model is generating entities closer to the distribution in the ground truth stories. We also conducted human evaluation that reveals that the glocal hierarchical attention model is preferred 82\% of the times. 

We plan on continuing the work mainly in the following three directions. Nominalized representation of the entity skeletons seem to outperform other models in METEOR score. We plan on investigating the incorporation of this skeleton form in the glocal attention model to reap the benefits of both the models. The second is that, despite the common usage of metrics such as METEOR for text generation tasks, it often lacks the needful targeted for the specific tasks. In our case, we have extended some analysis leveraging the number and the diversity of the entities generated. However, we plan on exploring metrics to evaluate intermediate tasks such as skeleton representation to streamline the end goal. Finally, we plan on applying our methods to other forms of conditions to generate storytelling such as semantic representations, graphs and prompts. This investigation paves way towards the generalizability of our approaches.


\bibliographystyle{aaai}
\bibliography{aaai}

\end{document}